\documentclass[a4paper,twoside]{article}

\usepackage{epsfig}
\usepackage{subcaption}
\usepackage{calc}
\usepackage{amssymb}
\usepackage{booktabs} 
\usepackage{tabularx}
\usepackage{array}
\usepackage{tabularx}
\usepackage{multirow}
\usepackage{multicol}
\usepackage{amstext}
\usepackage{amsmath}
\usepackage{adjustbox}
\usepackage{amsthm}
\usepackage{multicol}
\usepackage{pslatex}
\usepackage{apalike}
\usepackage{url}
\usepackage{siunitx}
\usepackage{pifont}
\newcommand{\etal}{et~al.}
\usepackage{algorithm2e}
\usepackage[bottom]{footmisc}
\usepackage{SCITEPRESS}     

\usepackage{amsmath,amsfonts,bm}

\def\eqref#1{equation~\ref{#1}}

\def\1{\bm{1}}

\def\vx{{\bm{x}}}

\def\vz{{\bm{z}}}

\def\mX{{\bm{X}}}

\DeclareMathAlphabet{\mathsfit}{\encodingdefault}{\sfdefault}{m}{sl}
\SetMathAlphabet{\mathsfit}{bold}{\encodingdefault}{\sfdefault}{bx}{n}

\def\sC{{\mathbb{C}}}

\newcommand{\Ctx}{\sC}
\newcommand{\sx}{s_{\vx}}
\newcommand{\xh}{\hat{\vx}}

\begin{document}

\title{Context-Aware Autoencoders for Anomaly Detection\\ in Maritime Surveillance}

\author{\authorname{Divya Acharya\sup{1}\orcidAuthor{0000-0003-0206-6428}, Pierre Bernabé\sup{1}\orcidAuthor{0000-0002-3567-8175}, Antoine Chevrot\sup{2}\orcidAuthor{0000-0003-3677-5150},\\ Helge Spieker\sup{1}\orcidAuthor{0000-0003-2494-4279}, Arnaud Gotlieb\sup{1}\orcidAuthor{0000-0002-8980-7585}, and Bruno Legeard\sup{2}\orcidAuthor{0000-0003-4986-7097}}
\affiliation{\sup{1}Simula Research Laboratory, Oslo, Norway}
\affiliation{\sup{2}Institut FEMTO-ST, Université de Bourgogne Franche-Comté, Besançon, France}
\email{\{divya,pierbernabe,helge,arnaud\}@simula.no, \{achevrot5,bruno\}@femto-st.fr}
}


\abstract{The detection of anomalies is crucial to ensuring the safety and security of maritime vessel traffic surveillance. Although autoencoders are popular for anomaly detection, their effectiveness in identifying collective and contextual anomalies is limited, especially in the maritime domain, where anomalies depend on vessel-specific contexts derived from self-reported AIS messages.
To address these limitations, we propose a novel solution: the context-aware autoencoder. By integrating context-specific thresholds, our method improves detection accuracy and reduces computational cost. We compare four context-aware autoencoder variants and a conventional autoencoder using a case study focused on fishing status anomalies in maritime surveillance.
Results demonstrate the significant impact of context on reconstruction loss and anomaly detection. The context-aware autoencoder outperforms others in detecting anomalies in time series data. By incorporating context-specific thresholds and recognizing the importance of context in anomaly detection, our approach offers a promising solution to improve accuracy in maritime vessel traffic surveillance systems.}

\onecolumn \maketitle \normalsize \setcounter{footnote}{0} \vfill

\section{\uppercase{Introduction}}

In open seas (beyond coastal areas), maritime surveillance is heavily based on detecting anomalous vessel behavior through the Automatic Identification System (AIS). AIS messages - self-declared by ships and collected by satellite - include time-stamped information on the identity, position, speed, and navigation status of a ship~\cite{international-maritime_organization-revised-2015}. With global maritime traffic on the rise, the volume of AIS messages exceeds 500 million per day. However, satellite-based AIS data is highly incomplete and unevenly distributed due to signal collisions, weather interference, and satellite orbit limitations.

Despite these limitations, anomaly detection in the AIS data remains crucial to identifying illicit activities, such as illegal fishing or status falsification. For example, two vessels of the same type (\textit{Set longliners}) with highly similar trajectories - yet one report being \textit{ on the way using the engine} and the other \textit{ involved in fishing}. The trajectory alone may not raise suspicion, but the mismatch between movement and reported status signals potential falsification. Traditional anomaly detection methods, unaware of this context, would fail to detect this discrepancy.

This research work is based on a large, real-world AIS dataset collected via polar-orbiting satellites. The dataset is unlabeled and manually inspected for behavior-based anomalies. We define anomalies as context-inconsistent behaviors, for example, a cargo vessel following a fishing-like pattern. To avoid data leakage, we split the trajectories by vessel ID over time. AIS context is defined as the combination of vessel type and navigational status, which can be self-reported or inferred.

Unsupervised anomaly detection is well suited to this domain, as labeled anomalies are rare and often subjective~\cite{Pang2021}, \cite{zhou2023self}, especially due to the lack of labeled data in maritime settings. Among unsupervised techniques, \textit{autoencoders} (AEs) are widely adopted. AEs learn a low-dimensional representation of input data and attempt to reconstruct it; unusually high reconstruction errors typically indicate deviations from learned patterns. However, conventional AEs treat all input data uniformly, using a shared decoder regardless of vessel type or status. This makes them unsuitable for detecting anomalies that are only apparent when viewed in a specific context, for example, a trajectory that is normal for a trawler (fishing boat) may be abnormal for a tanker (cargo vessel). Without modeling such a context explicitly, these anomalies remain undetected.

To address this, previous work has explored context-aware models. MoE-AE uses fully separate encoders and decoders per context, but becomes inefficient as the number of contexts grows. Chevrot et al.~\cite{Chevrot2022CAE} proposed the Contextual Autoencoder (CAE) with shared encoder and context-specific decoders, improving anomaly detection in avionics by modeling flight phases (climb, cruise, descent). However, maritime navigation lacks such clear-cut phases; vessel behavior is influenced by diverse types (e.g., trawler, tanker) and mission-specific status.
We propose to extend context-aware modeling to maritime anomaly detection using AE-based architectures. Our models incorporate context through dedicated decoders (CAE), fully separated branches (MoE-AE), or merged groups (GCAE). We explore the feasibility of context grouping to balance performance with computational efficiency. Here, “context” refers to a specific combination of vessel type and navigational status, known or inferred.

In this paper, we aim to advance the development of scalable context-aware autoencoders for real-world anomaly detection in maritime traffic surveillance and make the following three main contributions:
(1) We show the critical role of context awareness in detecting contextual anomalies and relevant collective anomalies when using autoencoders. (2) We propose a novel methodology called the Grouped Contextual Autoencoder (GCAE), which extends CAE and MoE-AE. GCAE allows for the identification and grouping of irrelevant contexts, reducing model size and complexity while maintaining good anomaly detection performance. (3) We compare the capability of four context-aware AEs (AE with context-specific thresholds, CAE, MoE-AE and GCAE) to detect collective and contextual anomalies against a regular AE in a maritime traffic surveillance case study.

\section{\uppercase{Related Work}}
\label{sec:related_work}

\subsection{Anomaly Detection}
Anomaly detection, also called outlier or novelty detection, identifies unusual patterns that significantly deviate from most previously seen data. It has applications in threat detection, cybersecurity, financial fraud, crowd analysis, and traffic surveillance~\cite{Chalapathy2019}.
As classified by \cite{Chandola2009}, anomalies are of three types: point, contextual, and collective. A \emph{point anomaly} is an individual outlier in relation to the rest of the data and is the simplest and most studied type. A \emph{contextual anomaly} (also called a conditional anomaly~\cite{4138201}) occurs when a data instance is anomalous only within a specific context. This relies on distinguishing \emph{behavioral attributes}, related to the data subject, from \emph{contextual attributes}, which are external factors influencing it~\cite{Shulman_19}. In spatio-temporal data, contextual anomalies break expected seasonality or spatial patterns and cannot be detected by point anomaly methods since their feature values remain within normal ranges. Lastly, a \emph{collective anomaly} is a related set of anomalous instances with respect to the entire dataset.

In time-series data, contextual anomalies manifest as irregular patterns, broken seasonality, or unexpected temporal changes. These cannot be detected by point anomaly methods, as feature values may remain within normal ranges. Traditional approaches often fail to capture such anomalies; most distance-based or clustering techniques, including DBSCAN~\cite{Ester1996} and IMS~\cite{Iverson2012}, focus on point anomalies and overlook temporal context. Recent IoT applications, such as HVAC anomaly detection~\cite{Zhang2022hvac}, have explored pattern-based methods like Dynamic Time Warping (DTW) to identify recurring or distorted temporal patterns, offering an alternative for context-aware detection.

Ensemble methods such as Isolation Forests~\cite{Liu2012}, even when combined with sliding windows~\cite{Ding2013} to capture temporal dependencies, struggle with context-aware anomaly detection. Some approaches transform time series into point anomalies~\cite{Ma2003} or detect repeated patterns using Dynamic Time Warping, as applied in HVAC systems~\cite{Zhang2022hvac}, though this fails when multiple samples cannot be separated an issue we observe in our experiments. A related class of methods trains models solely on normal data under regular patterns and flags deviations during inference. Examples include statistical forecasting models like ARIMA~\cite{Bianco2001}, used to predict future values, and short-term state prediction on power grids~\cite{Zhao2017}. Other variants apply deep neural networks~\cite{Yu2018} to detect abnormal deviations.

We refer to \cite{Pang2021} for a recent survey of learning techniques for anomaly detection. Perera et al.~\cite{perera2024ais} propose a spatio-temporal graph neural network approach for maritime vessel anomaly detection using AIS data. Their method effectively captures both spatial relationships and temporal dynamics, demonstrating improved detection accuracy over traditional techniques.

Prior approaches typically use a single decoder for all contexts, resulting in models trained on slightly different data distributions. While such models may achieve good reconstruction loss, they fail to capture context-specific variations—an essential aspect for accurate anomaly detection. Only a few works have explored context in this setting. Hayes \etal~\cite{Hayes2015} and Golmohammadi \etal~\cite{Golmohammadi2017} apply clustering-based context inference as a post-processing step to filter out false positives, rather than incorporating context during training.

\subsection{Anomaly Detection in AIS Data}
The detection of anomalies in the AIS data has gained interest in the literature since 2005 \cite{riveiro_maritime_2018}. 
Ribeiro et al.~\cite{ribeiro2023ais} provide a comprehensive review of AIS-based anomaly detection methods, emphasizing challenges such as data irregularities and real-time processing needs.
\cite{Pallotta_Vessel_2013}
Detection of specific anomaly
The recent works utilize more complex models but specialize in the detection of particular anomalies, making the model generalization difficult.
\cite{ford_detecting_2018} proposed a Generalized Additive Model (GAM), adapted to the open sea, which captures space-time variations in AIS gaps during transmissions to detect abnormal AIS missed reception.

Nguyen \etal \cite{Nguyen2019} propose an AIS message representation that regularizes the frequency of messages by completing the data set with artificially generated interpolated missing messages. 
The cleaned dataset can then be used in a multitask learning setting, or to learn a probabilistic map of vessel trajectories for anomaly detection purposes \cite{Nguyen2019}. Liang et al.~\cite{liang2024unsupervised} propose an unsupervised deep learning framework combining Wasserstein GANs and encoder networks to detect maritime anomalies without labeled data. Oh et al.~\cite{oh2025vae} compare various variational autoencoder monitoring statistics integrated with CUSUM, demonstrating improved real-time AIS anomaly detection performance.
Some approaches attempt to derive contextual importance directly from the data itself, without relying on predefined labels~\cite{fernandez_arguedas_unsupervised_2014}.

AEs have also been successfully applied for the detection of AIS anomalies. 
An AE has been shown to have better accuracy than support vector machines (SVM) or random forests in the classification of fishing activities \cite{Jiang2016}. 
More recently, vessel trajectory prediction has been addressed using one-dimensional convolutional AE \cite{Wang2021}. However, these approaches are not designed for context awareness.

\section{\uppercase{Method - Context-aware Autoencoders}}
\label{sec:context_aware_ae}
In the following, we introduce context-aware autoencoders and their extensions over the conventional AE.

\begin{figure*}[t]
    \includegraphics[width=\textwidth]{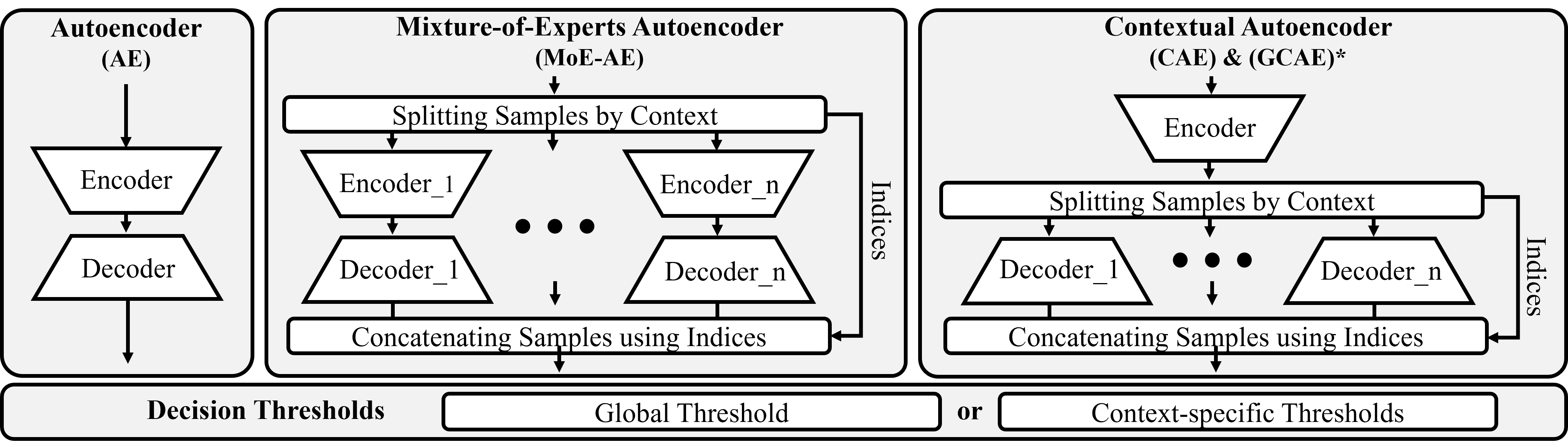}
    \caption{AE, MoE-AE, CAE \& GCAE architectures (*CAE has one decoder per context / GCAE has one per group of contexts)}
    \label{fig:models}
\end{figure*}       

All considered methods are end-to-end anomaly score learning methods that internally learn a feature representation of normality and identify anomalies through thresholds on the reconstruction loss.
An AE, adopting the notation from \cite{Pang2021}, consists of the encoder/decoder neural networks $\phi_{e}$ and $\phi_{d}$ with parameters $\Theta_{e}$ and $\Theta_{d}$, respectively.
The encoder encodes an input $\vx$ in the latent representation $\vz$, which is reconstructed again to $\xh$ by the decoder: $\hat{\vx} = \phi_{d}(\phi_{e}(\vx,\Theta_{e}),\Theta_{d})$.
The anomaly score $\sx$ for an input $\vx$ resembles directly its reconstruction loss $\sx = \| \vx - \xh\|^2$, and both the encoder and decoder are trained together to minimize this reconstruction error for the training set $\mX$. 

A \emph{context} is a categorical distinctor $c$ between $L$ data distributions $\Ctx = \{c_1, c_2, \dots, c_L\}$ relevant for an anomaly detection system. 
Distinguishing contexts is relevant when there is a distribution shift between contexts, i.e. the distribution of observations is similar for two contexts $c_i$ and $c_j$ ($P(\mX_{c_i}) = P(\mX_{c_j})$, but the distribution of what is an anomaly differs ($P(y_{c_i}) \neq P(y_{c_j})$).
Here, $y$ and $y_{c_i}$ are labels that indicate whether $\vx$ is a non-contextual ($y$) respectively a contextual anomaly in context $c_i$ ($y_{c_i}$): $y, y_{c_i} \in \{0,1\}$.
However, these labels do not exist during training, which is performed unsupervised with only the loss of reconstruction.
We also refer to the data subset belonging to a context $c \in \Ctx$ as $\mX_{c}$. 
The context for a data sample $\vx$ is given by $ctx(\vx) : \R^n \rightarrow \Ctx$.
This assignment can be made from a subset of the features in $\vx$, e.g., a status field, or can be determined externally.

\subsection{Conventional Autoencoder (AE)}
The conventional AE is capable of detecting \textit{ point anomalies} and \textit{collective anomalies}. However, it struggles with \textit{contextual anomalies} due to its inability to distinguish between different contexts. In essence, a conventional AE is designed to learn how to reconstruct all given data without any discernment. Even when providing contextual information as side input, it does not significantly improve the model's performance in anomaly detection. This is because the decision on whether a data point is an anomaly is not inherently built into the AE model.
Consequently, the AE strives to accurately reconstruct all inputs, regardless of their context. As a result, if certain behaviors are considered normal in one context but abnormal in another, the AE would not consider this discrepancy and will still reconstruct these behaviors accurately. This lack of context-sensitivity limits its effectiveness in detecting contextual anomalies.

\subsection{Mixture-of-Experts Autoencoder (MoE-AE)}
A naive approach to extending the AE to the contextual setting is to use a Mixture-of-Experts~\cite{Jacobs1991moe} autoencoder (MoE-AE).
The MoE-AE consists of one full AE per context $c$, that is, $|\Ctx|$ context-specific individual autoencoders, plus the additional routing to match inputs and the respective AE.
Their working and structure are the same as conventional AE, but they are trained exclusively on the data subset $\mX_{c}$ for their context.
During prediction time, the input is then routed to the corresponding AE, based on the context $ctx(\vx)$, to return the anomaly score $\sx$.
The advantage of this solution is that experts will never see any trajectory input specific to another context, unlike the conventional AE.
The disadvantage is that the size of the model multiplies with the number of contexts.
Additionally, it is harder to train on contexts with a small amount of data.

\subsection{Contextual Autoencoder (CAE)}
CAE~\cite{Chevrot2022CAE} is, like MoE-AE, context-aware and capable of detecting \textit{ contextual anomalies}.
In contrast to MoE-CAE, CAE has one joint encoder $\phi_{e}$ with parameters $\Theta_{e}$ for all contexts, while only decoders $\phi_{d,c}$ with parameters $\Theta_{d,c}$ are context-specific.
This follows from the intuition that even if the inputs are from different contexts, they are still from the same domain and share many of the same characteristics and patterns. Additionally, by sharing the encoder, we have more training samples available for it, which should strengthen the encoding and latent representation produced by the model.
However, the decoders are specialized to be able to reconstruct only a typical input belonging to their context, i.e., focusing on its specific characteristics in the overall domain distribution.
Finally, sharing the encoder also enables the expansion of the number of contexts by training only an additional decoder for the new context.

The joint encoder also has a cost advantage as it reduces the model by $(|\Ctx|-1)*|\Theta_{e,c_i}|$ parameters, that is, almost half of the total model size.

 \subsection{Grouped Contextual Autoencoder (GCAE) and Identification of Relevant Context}

To address the trade-off between model complexity and context-specific reconstruction, we propose a grouped contextual autoencoder (GCAE) architecture. This design maintains a single encoder but employs multiple decoder branches, each tailored to a group of contexts that share similar behavior.

Let $\mathcal{C} = \{c_1, c_2, \dots, c_n\}$ represent the set of context labels (defined by static or dynamic vessel attributes), and let $\mathcal{D}_c$ be the validation set corresponding to context $c \in \mathcal{C}$. Each decoder $d_i$ is trained in a subset of these contexts.

The grouping strategy is based on two core evaluations:
 \textbf{Contextual Anomaly Masking Test (CAMT):} Determines whether a decoder trained in context $i$ can reconstruct data from context $j$ without significant error. If for all $j \ne i$:
  \[
  \mathcal{L}(d_i, \mathcal{D}_j) > \mathcal{L}(d_j, \mathcal{D}_j) + \delta
  \]
  then context $j$ is considered unique and is not grouped. \newline
 \textbf{Decoder Irrelevance Test (DIT):} Identifies decoders that generalize in multiple contexts. A decoder $d_k$ is broadly applicable if:
  \[
  \forall j \in G \subseteq \mathcal{C},\quad \mathcal{L}(d_k, \mathcal{D}_j) \leq \tau
  \]
  where $\tau$ is a validation-derived loss threshold. All such contexts $G$ can then share the decoder $d_k$.

This formulation enables automatic grouping of indistinguishable contexts, reducing the total number of decoder branches while preserving anomaly sensitivity. The GCAE retains the contextual specificity of CAE while reducing the model complexity closer to that of a global AE.

\subsubsection{Identification of Relevant Contexts}
\label{sec:relevant_context_identification}

To determine whether distinct decoder branches are necessary for specific contexts, we introduce two quantitative tests: the Contextual Anomaly Masking Test (CAMT) and the Decoder Irrelevance Test (DIT). These tests enable systematic grouping of contexts while preserving anomaly discriminability.
Let $\mathcal{C}$ denote the set of identified contexts, and let $\mathcal{D}_c$ represent the validation set corresponding to context $c \in \mathcal{C}$. Let $d_i$ be a decoder trained exclusively in context $i$, and let $e(\cdot)$ be the shared encoder.
\newline
Contextual Anomaly Masking Test (CAMT):
This test evaluates whether the decoder $d_i$ trained in a different context $i \ne c$ performs significantly less compared to the dedicated decoder $d_c$ when reconstructing samples from context $c$. Specifically, we define the mean squared reconstruction loss as:
$
\mathcal{L}(d_j, \mathcal{D}_c) = \frac{1}{|\mathcal{D}_c|} \sum_{x \in \mathcal{D}_c} \left\|x - d_j(e(x))\right\|^2$
. The context $c$ is considered \textit{distinct} (requiring a dedicated decoder) if:
$\min_{j \neq c} \mathcal{L}(d_j, \mathcal{D}_c) > \mathcal{L}(d_c, \mathcal{D}_c) + \delta$
where $\delta > 0$ is a tolerance parameter that controls the sensitivity to reconstruction degradation.
\newline
Decoder Irrelevance Test (DIT):
In contrast, if a decoder $d_k$ generalizes well across multiple contexts, it can be shared. A context group $G \subseteq \mathcal{C}$ satisfies the DIT condition under decoder $d_k$ if:
$\forall c \in G,\quad \mathcal{L}(d_k, \mathcal{D}_c) \leq \tau$
where $\tau$ is a validation-derived loss threshold. This allows $\{ \mathcal{D}_c \mid c \in G \}$ to be reconstructed using a common decoder $d_k$.
These tests form the basis of our context grouping strategy, enabling the GCAE to balance compactness and sensitivity by merging only those contexts that are not statistically distinguishable in reconstruction behavior.

\subsection{Data Collection}
\label{sec:data_collection}

The AIS data in this study was provided by the Norwegian Coastal Administration and collected by Statsat AS via four satellites—AISSat-\{1,2\} and NorSat-\{1,2\}—operating in Sun-synchronous polar orbits at 600–650 km altitude, covering all latitudes.
The data set contains about 4.05 billion AIS messages recorded worldwide during 2020. Due to bandwidth and reception limits, the satellite AIS stream is incomplete; a similar volume was recorded along the Norwegian coast via terrestrial stations.
This incompleteness arises from (i) limited temporal/spatial sampling caused by orbital constraints, especially near the equator, and (ii) channel congestion in dense maritime regions (e.g., Europe, China). Furthermore, AIS was designed for ship-to-ship/shore communication, not satellite reception~\cite{skauen-ship-2019}.
The dataset is not publicly available due to ownership restrictions. A limited release covers the Norwegian economic zone\footnote{\url{https://www.kystverket.no/en/navigation-and-monitoring/ais/access-to-ais-data/}}, but there is no open-sea satellite AIS dataset of this scale publicly available. Community data sets such as AIShub\footnote{\url{https://www.aishub.net/}} mainly cover coastal areas, which are not suitable for the detection of open sea anomalies.

\subsection{Data Enrichment}

From each AIS message, we extract primary dynamic features: speed over ground ($s$), course over ground ($c$) and heading ($h$), augmented with derived features that capture temporal and spatial changes: time delta ($\Delta t$), geographic displacement ($\Delta d$) and bearing angle ($a$). This forms the feature vector $m = [s, c, h, \Delta t, \Delta d, a]$.
A trajectory $\mathcal{E}$ is a time-ordered sequence of AIS messages from the same vessel: $\mathcal{E} = [m_1, m_2, \dots, m_T]$.
To improve context discrimination, we enrich vessel type annotations by cross-referencing AIS vessel IDs with the Global Fishing Watch dataset,\footnote{\url{https://globalfishingwatch.org/data-download/datasets/public-fishing-effort}} covering over 114,000 fishing vessels classified by 16 fishing methods. We focus on the five most common types in our data: \textit{drifting longlines}, \textit{set longlines}, \textit{squid jiggers}, \textit{trawlers}, and \textit{tuna purse seines}, combined with navigational status to define context labels for anomaly detection.

\subsection{Dataset Creation}
\label{sec:dataset_creation}

\begin{figure}
    \centering
    \small
    \fbox{\begin{minipage}{\columnwidth}
\textbf{Under way using engine}: drifting longlines ($c_0$), set longlines ($c_1$), squid jigger ($c_2$), trawlers ($c_3$), tuna purse seines ($c_4$). \textbf{At anchor}: drifting longlines ($c_5$), trawlers ($c_6$), tuna purse seines ($c_7$) \textbf{Not under command}: drifting longlines ($c_8$). \textbf{Restricted maneuverability}: drifting longlines ($c_9$), squid jigger ($c_{10}$), trawlers ($c_{11}$). \textbf{Moored}: drifting longlines ($c_{12}$), squid jigger ($c_{13}$), trawlers ($c_{14}$), tuna purse seines ($c_{15}$). \textbf{Engaged in Fishing}: drifting longlines ($c_{16}$), set longlines ($c_{17}$), squid jigger ($c_{18}$), trawlers ($c_{19}$), tuna purse seines ($c_{20}$). \textbf{Under way sailing}: drifting longlines ($c_{21}$), set longlines ($c_{22}$), squid jigger ($c_{23}$), trawlers ($c_{24}$), tuna purse seines ($c_{25}$)
\end{minipage}}
\caption{Contexts in the dataset, each with associated vessel type.\label{fig:context_list}}
\end{figure}

Trajectories are assigned contexts based on vessel type and navigational status, focusing on the five main types of fishing vessels (Section~\ref{sec:data_collection}). Rare or sparse context pairs are discarded, leaving 25 unique context labels (see Figure~\ref{fig:context_list}).
To prevent data leakage, splits are made in the vessel MMSI. Each trajectory is a fixed sequence of 50 AIS messages, located at least 5 km from ports, to capture open-sea behavior.
Outlier trajectories with extreme time/distance gaps, poor coverage, or insufficient data are removed. To address class imbalance, dominant contexts are down-sampled (max 50,000 training, 5,000 validation/testing trajectories). The final dataset includes \num{1002633} training, \num{106582} validation, and \num{108220} testing trajectories.
The remaining imbalance is handled by sample weighting during training. Table~\ref{tab:dataset_stat} summarizes the statistics of the training set, reflecting the variability of satellite AIS sampling (Section~\ref{sec:data_collection}).

\begin{table}[t]
    \renewcommand{\arraystretch}{1.1}
    \centering
    \caption{Trajectory statistics for the training dataset after filtering and cleaning.}
    \adjustbox{max width=\columnwidth}{
    \begin{tabular}{lrrrrr}
        \toprule
        Percentile & 1st & 25th & 50th & 75th & 99th\\ \midrule        
        Dist. between messages [m] & 0 & 11 & 89 & 521 & 3556\\
        Time between messages [s]  & 1 & 11 & 32 & 216 & 9503\\
        Trajectory length [km]     & 0.07 & 8.9 & 24.2 & 48.5 & 136.9\\  
        Trajectory duration [h]    & 0.09 & 1.38 & 1.79 & 3.33 & 9.2 \\
        \bottomrule
    \end{tabular}}
    \label{tab:dataset_stat}
\end{table}

\subsection{Model Architectures}

All models evaluated in this study are based on an autoencoder (AE) framework composed of three components: an encoder, a decoder (or multiple decoders, depending on context handling), and a latent representation layer. Depending on the model variant, the encoder may be shared (AE, CAE, GCAE) or context-specific (MoE-AE).

The encoder and decoder are implemented using 1D convolutional neural networks (CNNs). The encoder consists of stacked \texttt{Conv1D}, \texttt{MaxPooling1D}, and \texttt{BatchNormalization} layers, while the decoder mirrors this structure using \texttt{Conv1DTranspose} and up sampling layers, as suggested in~\cite{yin_anomaly_2022}. The input to the model is a trajectory window of 50 messages, each containing 6 features (see Section~\ref{sec:data_collection}), resulting in an input tensor of shape $[50 \times 6]$ per sample.

To identify an optimal latent dimensionality, we empirically varied the size of the bottleneck layer while monitoring reconstruction fidelity. For each candidate latent size, encoder and decoder capacities were proportionally scaled to ensure convergence. When a further reduction in latent size caused a loss in reconstruction quality, we reverted to the previous configuration. This led us to select a latent representation size of 75 for the CNN-based models.
We also evaluated a model variant using LSTM layers~\cite{Hochreiter1997} in both encoder and decoder to better capture temporal dependencies. However, these recurrent models did not show a consistent advantage over the CNN-based architecture in terms of reconstruction quality or anomaly detection performance. Moreover, CNN-based models converged faster, showed lower variance across runs, and were more robust to hyperparameter settings. As such, all reported results are based on CNN architecture.

\begin{table}[t]
    \renewcommand{\arraystretch}{1.1}
    \centering
    \footnotesize
    \caption{Comparison of the proposed model variants.}
    \adjustbox{max width=\columnwidth}{
    \begin{tabular}{lcccc}
        \toprule
        Model & Encoder & Decoder(s) & Context-aware & Param. Count \\
        \midrule
        AE     & Shared & Shared         & \ding{55} & Low \\
        MoE-AE & Per context & Per context    & \ding{51}& High \\
        CAE    & Shared & Per context    & \ding{51} & Medium \\
        GCAE   & Shared & Grouped (by similarity) & \ding{51} & Low–Medium \\
        \bottomrule
    \end{tabular}}
    \label{tab:model_comparison}
\end{table}

\subsection{Training Setup}
\label{sec:training}

All models (AE, MoE-AE, CAE, GCAE) are trained for up to 250 epochs using the Adam optimizer with a learning rate of $1 \times 10^{-3}$ and a batch size of 128. The loss function is the mean squared error (MSE) between the input and the reconstructed trajectory.
To address the imbalance in context representation, especially in the GCAE setting, we apply sample weighting during training. This ensures that the shared encoder is not biased toward more frequent context types. Early stopping is used based on the validation reconstruction loss with a patience of 10 epochs, and the checkpoint achieving the best validation performance is used for evaluation.

Hyper parameters such as latent dimensionality were selected using a grid search on a held-out validation set. All models are implemented in PyTorch and executed on an NVIDIA DGX-2 system with an Intel Xeon Platinum 8168 CPU (2.7 GHz, 24 cores) and a single NVIDIA Tesla V100 GPU. Each experiment is repeated three times using different random seeds to account for variability.

\subsection{Thresholding and Anomaly Scoring}
\label{sec:thresholding}
During inference, each trajectory window $x$ is passed through the shared encoder and the appropriate decoder $d_c$ associated with its context $c$. The anomaly score is defined as the squared reconstruction error:
\begin{equation}
s(x) = \left\| x - d_c(e(x)) \right\|^2
\end{equation}
To determine whether $x$ is anomalous, we compare $s(x)$ with a context-specific threshold $\tau_c$, calculated from the training set $\mathcal{D}_c$. Assuming that reconstruction losses approximately follow a normal distribution, the threshold is defined as:
\begin{equation}
\tau_c = \mu_c + \lambda \cdot \sigma_c
\end{equation}
where $\mu_c$ is the mean loss of reconstruction in $\mathcal{D}_c$, $\sigma_c$ is the standard deviation of the reconstruction losses, and $\lambda$ is a scaling factor; we empirically use $\lambda = 5$.

This approach corresponds to a threshold $5\sigma$, chosen for its robustness in filtering out high-loss outliers while maintaining low false-positive rates. A sample $x$ is flagged as an anomaly if:
$s(x) > \tau_c$. 
Figure~\ref{fig:violin_plot_AE_test_collective_anoamlies}, presents the empirical distribution of validation reconstruction losses per decoder and context. The red dashed line indicates the corresponding threshold $\tau_c$. This visualization demonstrates that the majority of normal samples lie below the $5\sigma$ threshold, supporting its use for robust anomaly detection.

\subsection{Evaluation Metrics and Anomaly Matching}
We evaluated each model’s anomaly detection on open-sea vessel trajectories using unsupervised reconstruction loss (Section~\ref{sec:thresholding}), flagging segments exceeding context-specific thresholds $\tau_c$.
To compare AE, MoE-AE, CAE, and GCAE, we measured temporal overlap of flagged segments on the same trajectory, assessing coverage and agreement (Section~\ref{sec:discussion}).
Without labeled ground truth, we analyzed anomaly counts, contextual coverage, and detection overlap to assess model sensitivity, context discrimination, and decoder effectiveness.

\section{\uppercase{Experiments, Results, and Discussion}}
\label{sec:discussion}

This section presents our experimental results on the detection of AIS trajectory anomalies using four models based on autoencoders: a baseline AE with global reconstruction, a mixture-of-experts autoencoder (MoE-AE), a contextual autoencoder (CAE), and the proposed grouped contextual autoencoder (GCAE).
All experiments use the test dataset described in Section~\ref{sec:dataset_creation}, with anomalies identified using reconstruction loss and global or context-specific thresholds (Section~\ref{sec:thresholding}).

\subsection{Impact of the Context for Collective Anomalies}

Collective anomalies refer to groups of related data points that, while individually plausible, form anomalous patterns when viewed together. In this experiment, we examine how context-sensitive thresholding affects the detection of such anomalies.
We use a standard AE model and apply two thresholding strategies: a \textit{global threshold} computed from the entire validation dataset, and \textit{context-specific thresholds} computed independently for each context using the $5\sigma$ rule.

\begin{figure}[t]
    \centering
    \includegraphics[width=\columnwidth]{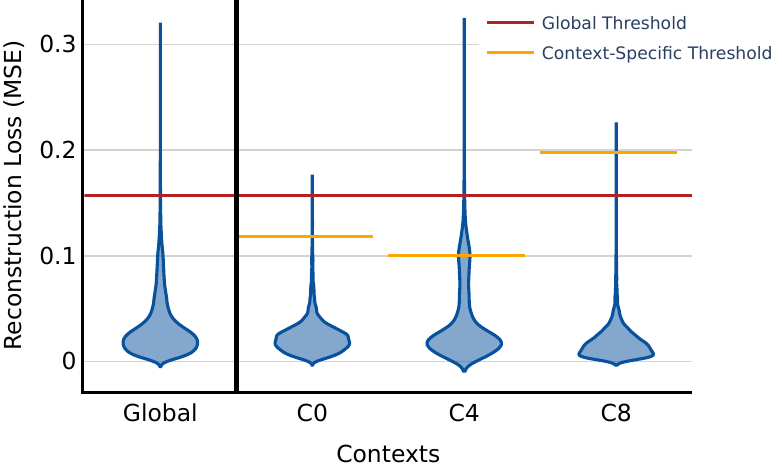}
    \caption{Validation reconstruction loss distributions per decoder and context. Each subplot corresponds to one decoder branch, with samples color-coded by context. The red dashed line represents the context-specific anomaly detection threshold $\tau_c$ computed using the $5\sigma$ rule. This highlights the intra-context variability and supports context-aware thresholding..}
    \label{fig:violin_plot_AE_test_collective_anoamlies}
\end{figure}

Figure~\ref{fig:violin_plot_AE_test_collective_anoamlies} presents the loss distributions for the AE reconstruction for three representative contexts in the test set: $c_0$, $c_4$ and $c_8$. The figure shows notable differences in threshold values. For example:
In context $c_0$ and $c_4$, the context-specific thresholds are 25\% and 36\% lower than the global threshold, respectively.
In contrast, for context $c_8$, the threshold is 25\% higher than the global one, almost twice as high as in $c_4$.
These variations are due to differences in reconstruction complexity. The contexts $c_0$ and $c_4$ involve vessels in straight-line motion (e.g., from point A to B), which are easier to model. In contrast, $c_8$ represents trawlers at anchor, whose movement patterns are irregular and driven by ocean currents, resulting in a higher loss of reconstruction and requiring a more lenient threshold.

Detection differences are quantified in Table~\ref{tab:confusion_collectives_anomalies}: the global threshold identifies 234 anomalies (0.22\% of the test set), whereas context-specific thresholds detect 292 (0.26\%). The overlap is poor despite similar totals: 136 anomalies (47\%) from context-specific thresholds are missed globally, and 78 (33\%) global anomalies are not flagged context-specifically. These results demonstrate that global thresholding introduces both false negatives and false positives due to its inability to adjust for context-dependent reconstruction variance, underscoring the importance of adapting thresholds to each context's behavioral characteristics

\begin{table}[t]
    \renewcommand{\arraystretch}{1.1}
    \centering
    \small
    \label{tab:confusion}
    \newcolumntype{C}{>{\centering\arraybackslash}X}%
    \begin{tabularx}{\columnwidth}{cCCCC}
        \toprule
        & & \multicolumn{2}{c}{Context-specific} & \multirow{2}{*}{Total}\\
        \cmidrule{3-4} 
        & & Normal & Anomaly &  \\ 
        \midrule
        \parbox[c]{6mm}{\multirow{2}{*}{\rotatebox[origin=c]{90}{\centering Global}}}
        & Normal & \num{107850} & \num{136} & \num{107986} \\
        & Anomaly & \num{78} & \num{156} & \num{234} \\
        \midrule
        \multicolumn{2}{c}{Total} & \num{107928} & \num{292} & \num{108220}\\
        \bottomrule
    \end{tabularx}
    \caption{Matrix of shared \textbf{collective} anomalies detected using global or context-specific thresholds on test dataset}
    \label{tab:confusion_collectives_anomalies}
\end{table}

\begin{figure}[t]
    \centering
        \includegraphics[width=\columnwidth]{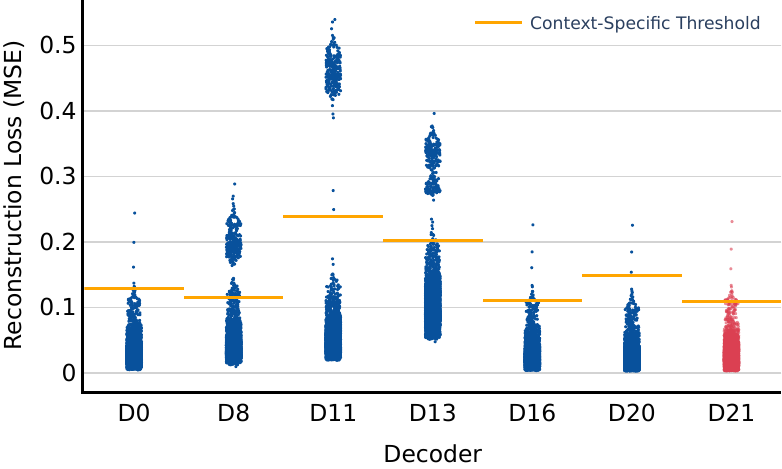}
    \caption{Test dataset reconstruction losses obtains with CAE on multiple decoders for context $c_{21}$ data (\textit{drifting longlines} vessels \textit{under way sailing}) through multiple decoders.}
    \label{fig:CAE_validation_violin_context_21_reduced}
\end{figure}

\subsection{Contextual Anomaly Detection}

A contextual anomaly refers to a data instance that is considered anomalous within a specific context but appears normal in other contexts. This differs from global anomalies, which are abnormal in all conditions. In maritime traffic surveillance, such context-specific deviations are particularly relevant due to the diversity of vessel types and navigational statuses.
We distinguish between two categories of attributes that contribute to contextual anomalies:
\textbf{Behavioral attributes:} These describe the physical motion of the vessel. For example, a vessel constrained by its draft that executes rapid course changes may be considered anomalous, even if similar trajectories are common for unconstrained vessels in other contexts.
\textbf{Contextual attributes:} These refer to inconsistencies between the reported metadata and the expected operational behavior. For example, a fishing vessel claiming to be \textit{under way using engine} while exhibiting patterns associated with \textit{engaged in fishing} activity violates contextual norms and raises suspicion.
Such anomalies cannot be reliably detected using global models, as the same trajectory can be deemed normal or abnormal depending on the declared type and status of the vessel. This reinforces the need for explicitly modeling context in anomaly detection systems, as proposed in our CAE and GCAE architectures.

\subsubsection{Undetected Anomalies in Conventional Autoencoders}

Figure~\ref{fig:CAE_validation_violin_context_21_reduced} illustrates the CAE reconstruction loss for the trajectories in the context $c_{21}$ (\textit{drifting longlines} vessels \textit{under way sailing}) when passed through various decoders. Decoder $D_{21}$, shown in red, was trained specifically on $c_{21}$ data and achieves the lowest reconstruction loss, as expected. Other decoders such as $D_0$, $D_{16}$, and $D_{20}$ can partially reconstruct this context due to behavioral similarity. However, several decoders — notably $D_8$, $D_{11}$, and $D_{13}$ — produce higher reconstruction loss, indicating that portions of $c_{21}$ data are not well represented outside their native context.
This demonstrates that certain behaviors are only accurately reconstructed when contextually modeled. A conventional AE, trained on all contexts jointly, would average these behaviors and potentially mask such differences.

To illustrate this limitation, consider a vessel that reports its status as \textit{under way sailing} (context $c_{21}$) while actually drifting with characteristics similar to \textit{not under command} (e.g., $c_8$). A global AE might fail to detect this deviation because it lacks context-specific expectations and thus minimizes loss across incompatible behaviors. In contrast, a CAE trained with separate decoders per context would detect such a mismatch due to the specialized representation learned for each context.
This confirms our hypothesis: explicit context modeling is critical for detecting anomalies that are only meaningful within their operational context.

\begin{table}[t]
    \renewcommand{\arraystretch}{1.0}
    \centering
    \small
    \newcolumntype{C}{>{\centering\arraybackslash}X}%
    \begin{tabularx}{\columnwidth}{cCCC}
        \toprule
        & \multicolumn{3}{c}{Detected anomalies} \\ 
        $\subseteq$ & AE & MoE-AE & CAE \\
        \midrule
        Total & 292 & 390 & 347\\
        \midrule
        AE & - & 230 (59\%) & 238 (69\%)\\
        MoE-AE & 230 (79\%) & - & 298 (86\%)\\
        CAE & 238 (82\%) & 298 (76\%) & - \\
        \bottomrule
    \end{tabularx}
    \caption{Comparison of anomalies detected by the models.}
    \label{tab:anomdetected}
\end{table}

\subsubsection{Context-aware Autoencoders}

Table~\ref{tab:anomdetected} summarizes the overlap in anomaly detection between AE, CAE, and MoE-AE on the test dataset. While AE identifies 292 anomalies, CAE and MoE-AE detect 347 and 390, respectively. Interestingly, both CAE and MoE-AE recover approximately 80\% of the anomalies found by AE. However, AE only recovers 59\% of the MoE-AE anomalies and 69\% of those detected by CAE.
This asymmetry highlights that context-aware models can detect a broader and more diverse set of anomalies—including subtle or context-specific deviations—missed by a conventional AE. These missed anomalies often arise from falsified metadata or behaviors that are only abnormal when viewed through the lens of a specific context.

A deeper inspection shows that not all discrepancies stem from model structure alone but also from threshold sensitivity. For example, in context $c_{21}$, the threshold for decoder $D_{21}$ (Figure~\ref{fig:CAE_validation_violin_context_21_reduced}) sits near the central mass of the distribution. This means that minor variations in loss can cause samples to switch classification around the threshold, leading to inconsistency in borderline cases.
To assess the severity of the anomaly, we computed the normalized distance of the detected samples above their respective thresholds. The 62 anomalies detected exclusively by AE are, on average, just 5\% above the threshold. In contrast, the 160 anomalies detected exclusively by MoE-AE are 71\% above threshold—indicating stronger evidence of abnormality.

We interpret this as a form of anomaly detection confidence: MoE-AE identifies more distinct and severe outliers, whereas AE tends to flag marginal cases that may be noise. CAE performs similarly to MoE-AE, though its mean distance to threshold is 17\%, slightly lower than the 23\% for MoE-AE. Despite this, CAE achieves comparable coverage with significantly reduced complexity—its parameter count is roughly half that of MoE-AE.
These results confirm that context-aware architectures like CAE and MoE-AE are more effective and precise than global AEs. Moreover, CAE offers a favorable trade-off between detection performance and model efficiency.

\subsection{Identification of the Most Relevant Context}
\label{sec:decoder_merging}

To reduce the complexity of the model while preserving detection quality, we investigate which context-specific decoders in the CAE can be combined. This analysis is guided by two conditions introduced in Section~\ref{sec:relevant_context_identification}: the \textit{Behavioral Condition} and the \textit{Contextual Condition}. Contexts satisfying both are considered redundant for anomaly modeling and can be grouped in the GCAE.
\par{\textbf{Behavioral Condition:}}
This condition verifies whether samples from a given context can be reconstructed by other decoders without being flagged as anomalous. For each context, we pass its data through all decoders and measure the percentage of samples incorrectly flagged as anomalies.
Context $c_{14}$ is poorly reconstructed by certain decoders, with up to 19.1\% of its samples flagged as anomalous, suggesting that its behavior is unique and context-specific. In contrast, contexts such as $c_{0}$ (0.68\%), $c_{5}$ (0.30\%), $c_{8}$ (0.02\%), $c_{12}$ (0.26\%), and $c_{16}$ (0.86\%) are consistently well reconstructed across decoders. These are considered to be behaviorally mergeable.
\par{\textbf{Contextual Condition}:}
This condition ensures that merging decoders will not mask important anomalies. We evaluate how well each decoder reconstructs samples from other contexts. A decoder is considered generalizable if it flags very few samples from other contexts as anomalous.
Decoder $D_8$, for instance, flags 5.52\% of other-context samples as anomalous—non-negligible leakage. However, several decoders exhibit very low cross-context anomaly rates: e.g., $D_4$ flags only 0.003\% of non-$c_4$ data. Filtering the behavioral candidates using this condition narrows the mergeable set to contexts $c_0$, $c_5$, $c_8$, $c_{12}$, and $c_{16}$. Among these, only $c_5$ satisfies both conditions completely.
\par{\textbf{Case-Specific Grouping Strategy:}}
In this case study, the primary focus is on detecting anomalies resulting from status falsification (e.g., a vessel claiming to be \textit{underway using engine} while exhibiting fishing behavior). This type of anomaly does not necessarily require behavioral uniqueness, and therefore we adopt a relaxed grouping strategy using only the Contextual Condition. Based on this, contexts $c_1$, $c_4$, $c_5$, $c_7$, $c_{10}$, and $c_{23}$—all showing less than 0.01\% false positives on other contexts—are grouped under a single decoder.
\par{\textbf{Effect of Grouping on Detection and Complexity}}
Applying the above strategy results in a GCAE composed of 20 decoders instead of the 25 used in CAE. The grouped contexts share a decoder, but thresholding remains context-specific. This architecture reduces the parameter count from \num{1964133} (CAE) to \num{1599824} (GCAE)—a 20\% reduction.
Despite the reduced complexity, detection performance remains stable. On the test set, CAE detects 347 anomalies and GCAE detects 352, with an 86\% overlap. The median difference in percentage of trajectories flagged as anomalies across decoders is just 0.29\%.
These results confirm that grouping by context simplifies the model architecture significantly without notably reducing anomaly detection performance.

\section{\uppercase{Conclusion}}
\label{sec:conclusion}

Detecting anomalies in the AIS trajectory data is challenging due to diverse vessel behaviors and limited satellite coverage. Although autoencoders are widely used for unsupervised detection, they often overlook contextual factors, reducing the relevance of detection.
We show that incorporating context, through vessel type and navigational status, significantly enhances the detection of collective and contextual anomalies. We compare four models: AE with context-specific thresholds, CAE, MoE-AE, and the proposed GCAE. MoE-AE offers the highest coverage, but at greater complexity, whereas CAE and GCAE provide strong performance with a reduced parameter count. GCAE, in particular, achieves a size reduction 20\% with 86\% anomaly overlap with CAE.
For resource-limited settings, CAE or GCAE are effective options. Even with standard AEs, the application of context-specific thresholds improves the relevance of the anomaly. This work provides a scalable, context-aware framework for maritime anomaly detection in real-world surveillance systems.

\section*{\uppercase{Acknowledgements}}
This work is funded by the European Commission through the project AI4COPSEC (Boosting EU Copernicus Security and Maritime Monitoring with AI and Machine Learning (ML), grant agreement No 101190021).

\bibliographystyle{apalike}
{\small
\bibliography{example}}

@article{Hochreiter1997,
  title = {Long Short-Term Memory},
  author = {Hochreiter, Sepp and Schmidhuber, Jurgen J{\"u}rgen},
  year = {1997},
  journal = {Neural Computation},
  volume = {9},
  number = {8}
}

@article{yin_anomaly_2022,
	title = {Anomaly Detection Based on Convolutional Recurrent Autoencoder for {IoT} Time Series},
	JOURNAL = {{IEEE} Transactions on Systems, Man, and Cybernetics: Systems},
	author = {Yin, Chunyong and Zhang, Sun and Wang, Jin and Xiong, Neal N.},
	year = {2022},

}

@techreport{international-maritime_organization-revised-2015,
	title = {{Revised} {guidelines} {for} {the} {onboard} {operational} {use} {of} {shipborne} {automatic} {identification} {systems} ({AIS})},
	pages = {19},
	number = {Resolution A.1106(29)},
	author = {{International Maritime Organization  ({IMO})}},
	institution = {IMO},
	year = {2015}
}

@article{riveiro_maritime_2018,
	title = {Maritime anomaly detection: A review},
	shorttitle = {Maritime anomaly detection},
	JOURNAL = {{WIREs} Data Mining and Knowledge Discovery},
	author = {Riveiro, Maria and Pallotta, Giuliana and Vespe, Michele},
	urldate = {2021-08-20},
	year = {2018},
}

@Article{Pallotta_Vessel_2013,
AUTHOR = {Pallotta, Giuliana and Vespe, Michele and Bryan, Karna},
TITLE = {Vessel Pattern Knowledge Discovery from {AIS} Data: A Framework for Anomaly Detection and Route Prediction},
JOURNAL = {Entropy},
VOLUME = {15},
YEAR = {2013},
NUMBER = {6},
PAGES = {2218--2245},
ISSN = {1099-4300}
}

@inproceedings{fernandez_arguedas_unsupervised_2014,
	year = {2014},
	author = {Fernandez Arguedas V and Pallotta G and Vespe M},
	title = {Unsupervised Maritime Pattern Analysis to Enhance Contextual Awareness },
    booktitle = {1st International Workshop on Context-Awareness in Geographic Information Services},
}

@article{Chandola2009,
author = {Chandola, Varun and Banerjee, Arindam and Kumar, Vipin},
title = {Anomaly Detection: A Survey},
year = {2009},
volume = {41},
number = {3},
issn = {0360-0300},
journal = {ACM Computing Surveys},
articleno = {15},
numpages = {58},
}

@article{Iverson2012,
author = {Iverson, David and Martin, Rodney and Schwabacher, Mark and Spirkovska, Lilly and Taylor, William and Mackey, Ryan and Castle, J.},
year = {2012},
month = {10},
title = {General Purpose Data-Driven System Monitoring for Space Operations},
volume = {9},
journal = {Journal of Aerospace Computing, Information, and Communication},
}

@article{Hayes2015,
  title={Contextual anomaly detection framework for big sensor data},
  author={Hayes, Michael A and Capretz, Miriam AM},
  journal={Journal of Big Data},
  year={2015},
}

@ARTICLE{Jacobs1991moe,
  author={Jacobs, Robert A. and Jordan, Michael I. and Nowlan, Steven J. and Hinton, Geoffrey E.},
  journal={Neural Computation}, 
  title={Adaptive Mixtures of Local Experts}, 
  year={1991},
}

@inproceedings{Golmohammadi2017,
  title={Sentiment analysis on twitter to improve time series contextual anomaly detection for detecting stock market manipulation},
  author={Golmohammadi, Koosha and Zaiane, Osmar R},
  booktitle={International Conference on Big Data Analytics and Knowledge Discovery},
  year={2017}
}

@article{chalapathy2019,
      title={Deep Learning for Anomaly Detection: A Survey}, 
      author={Raghavendra Chalapathy and Sanjay Chawla},
      year={2019},
      journal={arXiv:1901.03407}
}

@article{Ding2013,
  title={An anomaly detection approach based on isolation forest algorithm for streaming data using sliding window},
  author={Ding, Zhiguo and Fei, Minrui},
  journal={IFAC Proc. Vol.},
  volume={46},
  number={20},
  year={2013},
}

@article{Liu2012,
author = {Liu, Fei Tony and Ting, Kai Ming and Zhou, Zhi-Hua},
title = {Isolation-Based Anomaly Detection},
year = {2012},
journal = {ACMTKDD},
}

@inproceedings{Ester1996,
author = {Ester, Martin and Kriegel, Hans-Peter and Sander, J\"{o}rg and Xu, Xiaowei},
title = {A Density-Based Algorithm for Discovering Clusters in Large Spatial Databases with Noise},
year = {1996},
booktitle = {Second International Conference on Knowledge Discovery and Data Mining},
pages = {226–231},
series = {KDD'96}
}

@article{Bianco2001,
author = {Bianco, Ana and Garcia Ben, Marta and Martinez, Eunie Jr and Yohai, Victor},
year = {2001},
title = {Outlier Detection in Regression Models with ARIMA Errors using Robust Estimates},
volume = {20},
journal = {Journal of Forecasting}
}

@INPROCEEDINGS{Ma2003,
  author={Ma, J. and Perkins, S.},
  booktitle={{IJCNN}}, 
  title={Time-series novelty detection using one-class support vector machines}, 
  year={2003},
}

@ARTICLE{Yu2018,
  author={Yu, James J. Q. and Hou, Yunhe and Li, Victor O. K.},
  journal={IEEE Transactions on Industrial Informatics}, 
  title={Online False Data Injection Attack Detection With Wavelet Transform and Deep Neural Networks}, 
  year={2018},
}

@ARTICLE{Zhao2017,
  author={Zhao, Junbo and Zhang, Gexiang and La Scala, Massimo and Dong, Zhao Yang and Chen, Chen and Wang, Jianhui}, journal={IEEE Transactions on Smart Grid},
  title={Short-Term State Forecasting-Aided Method for Detection of Smart Grid General False Data Injection Attacks}, year={2017},
  volume={8},
  number={4},
  pages={1580-1590},
}

@article{Nguyen2019,
  title = {GeoTrackNet-A Maritime Anomaly Detector Using Probabilistic Neural Network Representation of AIS Tracks and A Contrario Detection},
  author = {Nguyen, Duong and Vadaine, Rodolphe and Hajduch, Guillaume and Garello, Ren{\'e} and Fablet, Ronan},
  year = {2021},
  journal = {IEEE Transactions on Intelligent Transportation Systems}
}

@article{skauen-ship-2019,
	title = {Ship tracking results from state-of-the-art space-based {AIS} receiver systems for maritime surveillance},
	journal = {{CEAS} Space Jrnl},
	author = {Skauen, Andreas Nordmo},
	year = {2019}
}

@article{ford_detecting_2018,
	title = {Detecting suspicious activities at sea based on anomalies in Automatic Identification Systems transmissions},
	journal = {{PLoS} {ONE}},
	author = {Ford, Jessica H. and Peel, David and Kroodsma, David and Hardesty, Britta Denise and Rosebrock, Uwe and Wilcox, Chris},
	year = {2018},
}

@article{Chevrot2022CAE,
title = {{CAE}: Contextual auto-encoder for multivariate time-series anomaly detection in air transportation},
journal = {Computers \& Security},
volume = {116},
year = {2022},
author = {Antoine Chevrot and Alexandre Vernotte and Bruno Legeard},
}

@ARTICLE{4138201,
  author={Song, Xiuyao and Wu, Mingxi and Jermaine, Christopher and Ranka, Sanjay},
  journal={IEEE Transactions on Knowledge and Data Engineering}, 
  title={Conditional Anomaly Detection}, 
  year={2007},
  volume={19},
  number={5},
  pages={631-645}
}

@inproceedings{Wang2021,
  title={AIS ship trajectory clustering based on convolutional auto-encoder},
  author={Wang, Taizheng and Ye, Chunyang and Zhou, Hui and Ou, Mingwang and Cheng, Bo},
  booktitle={Proceedings of SAI Intelligent Systems Conference},
  pages={529--546},
  year={2020},
  organization={Springer}
}

@InProceedings{Jiang2016,
author="Jiang, Xiang
and Silver, Daniel L.
and Hu, Baifan
and de Souza, Erico N.
and Matwin, Stan",
title="Fishing Activity Detection from AIS Data Using Autoencoders",
booktitle="Advances in Artificial Intelligence",
year="2016",
}

@article{Shulman_19,
  title={Unsupervised contextual anomaly detection using joint deep variational generative models},
  author={Shulman, Yaniv},
  journal={arXiv:1904.00548},
  year={2019}
}

@inproceedings{zhou2023self,
  title={Self-supervised Learning for Time-Series Anomaly Detection using Transformers},
  author={Zhou, Tianlin and Li, Yuhan and Huang, Sheng},
  booktitle={ICLR},
  year={2023}
}

@article{perera2024ais,
  title={AIS-based Maritime Vessel Anomaly Detection using Spatio-Temporal Graph Neural Networks},
  author={Perera, Chamath and Zhang, Wei and Lee, John},
  journal={IEEE Transactions on Intelligent Transportation Systems},
  year={2024},
  volume={25},
  number={3},
  pages={2334--2346}
}

@article{oh2025vae,
  author    = {YongKyung Oh and Kwonin Yoon and Jaemin Park and Sungil Kim},
  title     = {Comparative evaluation of VAE-based monitoring statistics for real-time anomaly detection in AIS data},
  journal   = {Maritime Policy \& Management},
  volume    = {52},
  number    = {4},
  pages     = {609--626},
  year      = {2025},
  publisher = {Taylor \& Francis},
  doi       = {10.1080/03088839.2024.2388177},
  url       = {https://doi.org/10.1080/03088839.2024.2388177}
}

@article{liang2024unsupervised,
  author    = {Maohan Liang and Lingxuan Weng and Ruobin Gao and Yan Li and Liang Du},
  title     = {Unsupervised maritime anomaly detection for intelligent situational awareness using AIS data},
  journal   = {Knowledge-Based Systems},
  volume    = {284},
  pages     = {111313},
  year      = {2024},
  publisher = {Elsevier},
  doi       = {10.1016/j.knosys.2023.111313},
  url       = {https://doi.org/10.1016/j.knosys.2023.111313}
}

@article{ribeiro2023ais,
  author    = {Cláudio V. Ribeiro and Aline Paes and Daniel de Oliveira},
  title     = {AIS-based maritime anomaly traffic detection: A review},
  journal   = {Expert Systems with Applications},
  volume    = {231},
  pages     = {120561},
  year      = {2023},
  publisher = {Elsevier},
  doi       = {10.1016/j.eswa.2023.120561},
  url       = {https://doi.org/10.1016/j.eswa.2023.120561}
}

@article{Zhang2022hvac,
  author={Zhang, Rui and Chen, Jian and Lin, Yiyang and Li, Xiaorui},
  title={Pattern-Based Contextual Anomaly Detection in HVAC Systems Using Time Series Shape Similarity},
  journal={IEEE Internet of Things Journal},
  year={2022},
  volume={9},
  number={4},
  pages={2690--2702},
  doi={10.1109/JIOT.2021.3098437}
}

@article{Pang2021,
author = {Pang, Guansong and Shen, Chunhua and Cao, Longbing and Hengel, Anton Van Den},
title = {Deep Learning for Anomaly Detection: A Review},
year = {2021},
volume = {54},
number = {2},
journal = {ACM Comput. Surv.},
articleno = {38}
}



\end{document}